# AIONER: All-in-one scheme-based biomedical named entity recognition using deep learning

Ling Luo[1,2,+], Chih-Hsuan Wei[1,+], Po-Ting Lai[1], Robert Leaman[1], Qingyu Chen[1] and Zhiyong Lu[1,*]

[1]National Center for Biotechnology Information (NCBI), National Library of Medicine (NLM), National Institutes of Health (NIH), Bethesda, MD 20894, USA

[2]School of Computer Science and Technology, Dalian University of Technology, Dalian 116024, China

[*]To whom correspondence should be addressed. Email: zhiyong.lu@nih.gov

## Abstract

Biomedical named entity recognition (BioNER) seeks to automatically recognize biomedical entities in natural language text, serving as a necessary foundation for downstream text mining tasks and applications such as information extraction and question answering. Manually labeling training data for the BioNER task is costly, however, due to the significant domain expertise required for accurate annotation. The resulting data scarcity causes current BioNER approaches to be prone to overfitting, to suffer from limited generalizability, and to address a single entity type at a time (e.g., gene or disease). We therefore propose a novel all-in-one (AIO) scheme that uses external data from existing annotated resources to enhance the accuracy and stability of BioNER models. We further present AIONER, a general-purpose BioNER tool based on cutting-edge deep learning and our AIO schema. We evaluate AIONER on 14 BioNER benchmark tasks and show that AIONER is effective, robust, and compares favorably to other state-of-the-art approaches such as multi-task learning. We further demonstrate the practical utility of AIONER in three independent tasks to recognize entity types not previously seen in training data, as well as the advantages of AIONER over existing methods for processing biomedical text at a large scale (e.g., the entire PubMed data).

Keywords: Biomedical named entity recognition; All-in-one; Deep learning; Natural language processing

## 1 Introduction

Large-scale application of automated natural language processing (NLP) to biomedical text has successfully helped address the information overload resulting from the thousands of articles added to the biomedical literature daily (Sayers *et al.*, 2021). Biomedical NLP is also increasingly used to support quantitative biomedical science, by augmenting manual curation efforts to populate databases via automated extraction (Singhal *et al.*, 2016), or by making inferences directly, through tasks such as literature-based knowledge discovery (Weeber *et al.*, 2001). Biomedical named entity recognition (BioNER), the task of identifying bio-entities such as chemicals and diseases within the text, provides an important foundation for many biomedical NLP

---

[+] The authors wish it to be known that, in their opinion, the first two authors should be regarded as joint first authors.

applications, and the accuracy of the entities identified by BioNER strongly affects the quality of the downstream applications. However, biomedical entities are with much more complicated naming principles. Compared with NER tasks in the general domain, such as the recognition of the persons or organizations, BioNER is more challenging because biomedical entity names are longer, more complex, and ambiguous (Cariello *et al.*, 2021; Jeong and Kang, 2021).

Prior to the deep learning era, conditional random fields (CRF) (Lafferty *et al.*, 2001) with the rich feature sets were the most popular method for BioNER, and consistently performed well on a variety of tasks (Leaman *et al.*, 2013; Leaman *et al.*, 2015; Wei *et al.*, 2015). In recent years, several deep learning-based methods have been widely applied to BioNER tasks with promising results, including bidirectional Long Short-Term Memory with a CRF layer (BiLSTM-CRF) (Lample *et al.*, 2016), Embeddings from Language Models (ELMo) (Peters *et al.*, 2018), and Bidirectional Encoder Representations from Transformers (BERT) (Devlin *et al.*, 2019). Most recently, the BERT pre-trained language model has become among the most popular methods, and several variants trained on biomedical text have been publicly released and widely applied to BioNER tasks (e.g., BlueBERT (Peng *et al.*, 2019), BioBERT (Lee *et al.*, 2020), and PubMedBERT (Gu *et al.*, 2021)).

The success of these machine-learning based methods relies heavily on manually annotated gold-standard data for model training and testing. Hence, significant efforts have been made to develop BioNER corpora for key biomedical entities such as diseases (Doğan *et al.*, 2014) and chemicals (Krallinger *et al.*, 2015). However, unlike the general English domain, manually annotating biomedical text requires domain knowledge and is highly costly. As a result, the current corpora in BioNER are generally limited in size, with a few hundred articles on average, and machine learning models trained on such limited annotations are prone to overfitting. Several recent studies (Galea *et al.*, 2018; Langnickel and Fluck, 2022) demonstrate that the accuracy of models trained on individual corpora decreases significantly when applied to independent corpora due to the limited generalizability of entity-related features captured by individual corpora.

Recently, several BioNER methods based on multi-task learning (MTL) (Crichton *et al.*, 2017; Giorgi and Bader, 2020; Rodriguez *et al.*, 2022; Wang *et al.*, 2019; Zuo and Zhang, 2020) have been proposed, which improve the generalizability of the model by making use of various publicly available datasets. These methods generally share the hidden layers of the deep learning model across the related tasks and have an output layer specific to each task. In particular, MTL can improve the model's generalizability by leveraging domain-specific information found in the training signals of related tasks (Caruana, 1997). However, several studies (Chai *et al.*, 2022; Rodriguez *et al.*, 2022) have found that MTL results are not always stable: MTL can improve performance compared to single-task learning on some datasets, but does not do so universally. Moreover, the MTL-based methods (Chai *et al.*, 2022; Wang *et al.*, 2019) usually require a complex model architecture.

In this work, we propose a novel data-centric perspective to enhance the accuracy and robustness of BioNER models by merging multiple datasets into a single task via task-oriented tagging labels. As a result, our method can achieve better performance more consistently than MTL and is applicable to various machine learning models.

More specifically, we propose AIONER, a new BioNER tagger that takes full advantage of various existing datasets for recognizing multiple entities simultaneously, despite their inherent differences in scope and quality, through a novel all-in-one (AIO) scheme. Our AIO scheme utilizes a small dataset recently annotated with multiple entity types (e.g., BioRED (Luo *et al.*, 2022a)) as a bridge to integrate multiple datasets annotated with a subset of entity types, thereby recognizing multiple entities at once, resulting in improved accuracy and robustness. Experimental results show that external datasets can help AIONER achieve statistically higher performance compared to that of the model trained only using the original BioRED training data. Vice versa, we demonstrate our AIO scheme can help improve model performance on individual datasets.

We further show that instead of using the entire BioRED training data (500 articles) in AIONER, we can use a minimal set of 10 articles to achieve a competitive result (86.91%) when external datasets are used. Finally, we demonstrate that AIONER can be reused in a versatile manner as a pre-trained model to further improve the performance of other BioNER tasks, even when the entity types are not previously seen in the AIONER training data.

## 2 Methods

The overall architecture of AIONER for multiple named entity recognition is shown in Figure 1. We first collected multiple resources for the six target entity types (i.e., gene, disease, chemical, species, variant, and cell line) which were annotated in the BioRED dataset. We then propose an effective all-in-one strategy to merge different resources into a single sequence labeling task. Next, a cutting-edge deep learning model is trained with the merged dataset for this BioNER task. Finally, the trained model is used to recognize the multiple biomedical entities from unseen documents. Further details on each step are provided in the following subsection.

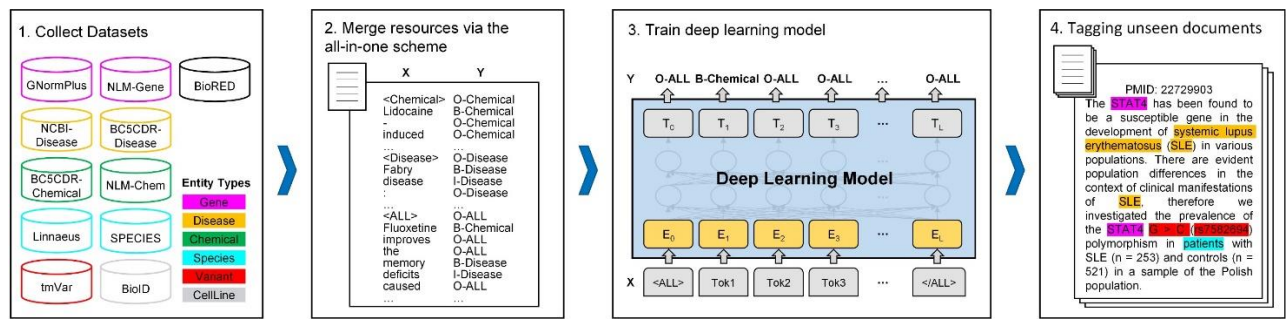

**Fig. 1.** Overview of our AIONER pipeline.

### 2.1 Publicly available BioNER datasets

To develop such a comprehensive BioNER method, we collected multiple resources within the six most popular entities in biomedical literature (i.e., gene, disease, chemical, species, variant, and cell line) as shown in Supplementary Table 1. We defined two criteria to filter the inconsistent datasets: (1) The selected datasets should annotate the interchangeable entities consistently. Some concepts are highly relevant and are usually used interchangeably. For example, a drug is a chemical substance that affects the functioning of living things and is frequently represented by a chemical. Besides, the gene and its products (e.g., RNA and protein) are usually named identically. There are also some other overlapping concepts, like phenotypes to diseases and residues to variants. (2) The datasets should annotate the concept identifiers of the entities by the same resources (e.g., NCBI Gene for gene/protein and MESH for chemical), which guarantees the definitions of the curated entities are consistent. The narrowed down resources are shown in Table 1.

**Table 1.** The BioNER datasets used in our study

| Entity type | Dataset | Text Size | Entities |
| --- | --- | --- | --- |
| All | BioRED(Luo *et al.*, 2022a) | 600 abs | 20,419 |
| Gene | GNormPlus (Wei *et al.*, 2015) | 694 abs | 9,986 |
|  | NLM-Gene (Islamaj *et al.*, 2021b) | 550 abs | 15,553 |
| Disease | NCBI Disease (Doğan *et al.*, 2014) | 793 abs | 6,892 |
|  | BC5CDR-Disease (Li *et al.*, 2016) | 1,500 abs | 12,850 |
| Chemical | BC5CDR-Chemical (Li *et al.*, 2016) | 1,500 abs | 15,935 |
|  | NLM-Chem (Islamaj *et al.*, 2021a) | 150 full | 40,467 |
| Species | Linnaeus (Gerner *et al.*, 2010) | 100 full | 4,259 |
|  | Species-800 (Pafilis *et al.*, 2013) | 800 abs | 3,708 |
| Variant | tmVar3 (Wei *et al.*, 2022) | 500 abs | 1,895 |
| Cell line | BioID (Arighi *et al.*, 2017) | 570 full | 5,590 |

Abs denotes abstracts; full denotes full-texts. Text genre is scientific article.

However, a few minor inconsistencies remained in the selected corpora after filtering. First, the annotations in the BioID (Arighi *et al.*, 2017) dataset are inconsistent internally. About 30% of the cell line spans with the suffix "cell(s)". To make it more consistent, we removed all the suffixes before the training and evaluation. Besides, Linnaeus (Gerner *et al.*, 2010) and Species-800 (Pafilis *et al.*, 2013) do not annotate the species relevant clinical terms (e.g., patient). Also, Species-800 excludes the genus name (e.g., Arabidopsis), and the species name which is in the higher level of the taxonomy system (e.g., Fungal), although those are annotated in BioRED. Third, GNormPlus (Wei *et al.*, 2015) and NLM-Gene (Islamaj *et al.*, 2021b) distinguish gene/protein family (e.g., Dilps) from the specific gene names (e.g., Dilp6), but BioRED did not distinguish the two types of entities. In our implementation, we merged the family names to the gene entity type.

## 2.2 All-in-one scheme

Like most previous studies, we treated BioNER as a sequence labeling task. Consider a sequence of text $\mathbf{X} = (x_1, x_2, \ldots, x_n)$, where $n$ denotes the length of the text. Each $x$ is tagged with a class label $y \in \mathbf{Y}$, where $\mathbf{Y}$ denotes the tagging scheme set (e.g., BIO scheme). The BIO scheme (Sang and De Meulder, 2003), which contains begin tokens ("B"), inside tokens ("I"), and background (outside) tokens ("O"), is the most popular encoding scheme of the BioNER task. Unlike the traditional BIO scheme, we designed and proposed a novel all-in-one (AIO) scheme to accept multiple datasets from different tasks. Specifically, given $m$ tasks and considering an input sentence $\mathbf{X}$ from the task $T_i$ where $i \in \{1, \ldots, m\}$, we applied an additional pair of tags surrounding the input sentence to indicate the task $\mathbf{X} = (<Task_i>, x_1, x_2, \ldots, x_n, </Task_i>)$ (e.g., "<Disease></Disease>" to recognize disease entities, and "<ALL></ALL>" to recognize all concept entities). The special tokens indicating the task tags were added to the beginning and end of the input sentence. For the label set **Y**, we defined 3 types of labels, which include "B-EntityType", "I-EntityType", and "O-Task". Note that, the definitions of the "B" and "I" in AIO scheme are the same as the traditional BIO scheme. However, to avoid entity conflict, we flexibly redesigned the "O" (outside) label since some entities may be curated in some of the datasets but not others. For example, in the scenario of recognizing diseases, the original "O" label is modified to "O-Disease" which can be clearly distinguished from the "O-Chemical" label for the task of recognizing chemical entities. Finally, we defined a total of 19 labels in the label set **Y** = {B-Gene, I-Gene, O-Gene, B-Disease, I-Disease, O-Disease, …, B-CellLine, I-CellLine, O- CellLine, O-ALL}.

We merged the datasets listed in Table 1 based on the AIO scheme. Excluding BioRED annotating all entity types, other datasets only focus on partially annotated entity types. If the dataset contains multiple partially annotated entity types, we split it into multiple datasets with a single entity type (e.g., BC5CDR is split into BC5CDR-Disease and BC5CDR-Chemical). Then we merged the datasets with the same entity type into a BioNER task. Figure 2 shows some example sentences annotated based on our tagging scheme. For example, we merged GNormPlus and NLM-Gene datasets for the gene recognition task. All sentences in these two datasets are added "<Gene>" and "</Gene>" tags at the front and end of the sentence. Only the tokens of the gene entity are labeled as "B-Gene" (or "I-Gene"). All other tokens are labeled as "O-Gene". "B-Gene" represents the status of the first token of the gene span, "I-Gene" represents the tokens of the gene span other than the first one, "O-Gene" represents the background tokens out of the gene spans. Different from other datasets, BioRED is a resource containing all entity types. We added "<ALL></ALL>" tags surrounding the input sentence from BioRED to indicate recognizing all entities of the six entity types. The tokens of the biomedical entity are labeled to B-EntityType (or I-EntityType), and all background tokens are labeled with "O-ALL". After converting all datasets using the AIO scheme, the integrated data is used to train NER models.

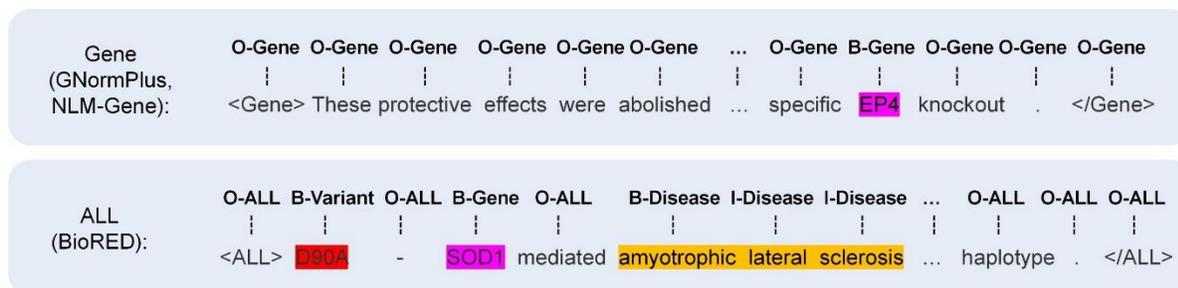

**Fig. 2.** Some example sentences annotated based on our AIO scheme.

**2.3 Deep learning model for BioNER**

The BioNER task has been modulated to a sequence labeling task via the AIO scheme. We applied cutting-edge biomedical pre-trained language models (PLMs) (e.g., PubmedBERT (Gu *et al.*, 2021)) for the implementation of our framework. Specifically, given an input sentence $\mathbf{X} = (x_1, x_2, …, x_n)$ consisting of $n$ tokens, the model aims to map the token sequence to a corresponding sequence of the label $\mathbf{y} = (y_1, y_2, …, y_n)$, where $y \in \mathbf{Y}$. We first used the PLMs to encode the input to a hidden state vector sequence, then computed the network score using a fully connected layer with a ReLU (Glorot *et al.*, 2011) activation. Finally, the CRF output layer is added to optimize the boundary detection of the bio-entities. A score is defined as:

$$s(\mathbf{X}, \mathbf{y}) = \sum_{i=1}^{n} \left( T_{y_{i-1}, y_i} + P_{i, y_i} \right) \tag{1}$$

Where **P** is the score matrix of the output from the last fully connected layer and **T** is a transition matrix of the CRF layer. $T_{i,j}$ represents the score of the transition from the ith label to the jth label. During the training phase, the objective of the model is to maximize the log-probability of the correct tag sequence:

$$log(p(\mathbf{y}|\mathbf{X})) = log\left( \frac{e^{s(\mathbf{X}, \mathbf{y})}}{\sum_{\tilde{\mathbf{y}}} e^{s(\mathbf{X}, \tilde{\mathbf{y}})}} \right) \tag{2}$$

Where $\tilde{\mathbf{y}}$ denotes all possible tag paths. At inference time, we predict the tag path that obtains the maximum score given by:

$$y = arg\, max_{\tilde{\mathbf{y}}}\, s(\mathbf{X}, \tilde{\mathbf{y}}) \tag{3}$$

This can be computed using dynamic programming, and the Viterbi algorithm (Viterbi, 1967) has been chosen for this inference.

We merged the training and development sets of the datasets for model training and evaluated the models on the official test sets. Since the official test sets of Linnaeus, Species-800 and BioID are not available or released, we randomly split 20% of the dataset as the test set for evaluating the models. We applied the default PLM parameter settings and set main hyper-parameters as follows: learning rate of 5e-6, batch size of 32, and max input length of 256 tokens (split into multiple sentences if the length of a sentence is over the max length). To determine the optimal number of training epochs, we set the patience parameter to 5 with a maximum of 50 epochs. The training process would terminate if there were no significant accuracy improvements in five consecutive epochs. For a detailed account of hyper-parameter settings, please refer to the Supplementary Material.

After training the model, the trained model can be used for recognizing the biomedical entities from the unseen input text. First, the input text was split into sentences and tokenized. Then we inserted the special task tags surrounding the input sentence to indicate the task. Finally, the trained AIONER model can be used to tag the tokens of the sentence to extract the task-specific entities according to the inserted task tags (e.g., "<Disease></Disease>" to only recognize disease entities, and "<ALL></ALL>" to recognize all concept entities). It should be noted that the entity scope and definitions in individual corpora may not align completely with BioRED. Therefore, utilizing the "<ALL> </ALL>" task tags for identifying all concept entities in Bi-oRED and applying individual (IND) task tags for identifying entities in the corresponding individual corpora. Further information can be found in Table S3 of the Supplementary Material.

## 3 Results

### 3.1 Experimental settings

To demonstrate the effectiveness of AIONER, we performed four experiments. First, we examined the performance of AIONER for multiple entity recognition on the BioRED test set. Second, we evaluated the stability and robustness of AIONER by analyzing its overall performance on the test set for each individual dataset. Third, we tested whether AIONER can be applied to support those BioNER tasks with new entity types that are not previously seen during AIONER model training. Finally, we investigated the performance and efficiency of the different variants of the BERT-based pre-trained model in our framework for supporting the processing of PubMed-scale literature data.

Additionally, we also implemented the MTL framework with the same deep learning model and training data for comparison. MTL treats each dataset as an individual task, and its model architecture shares the hidden layers across the different tasks where each task has its own task-specific output layers. The final loss is calculated by summarizing the losses of different tasks. In our experiments, we evaluated the model performance using the entity-level micro F1-score (F1), which has been widely applied in BioNER tasks. We further applied the two-sided Wilcoxon signed-rank test to perform statistical significance testing. Note that a few documents exist in both BioRED and some other datasets. To accurately evaluate the performance of different methods, we filtered those overlapped documents in the training set if the documents also exist in the test set.

**3.2 Multiple named entity recognition via AIONER on BioRED**

We examined the effectiveness of the AIO scheme and the contribution of different datasets for multiple named entity recognition. According to the evaluation in a 2022 study (Luo *et al.*, 2022), the PubMedBERT-CRF model achieves state-of-the-art (SOTA) performance and compares favorably to other methods such as BiLSTM-CRF and BioBERT-CRF models on the BioRED dataset. Therefore, we use it as the default model in our architecture of AIONER and MTL. We firstly prepared a strong baseline based on the PubMedBERT-CRF model, which trained on the original BioRED training set. Then we merged every external dataset and BioRED training set via our AIO scheme to train the models, respectively. Finally, we integrated all datasets as a union training set and trained the AIONER model (i.e., BioRED+All (AIONER)). We also implemented two more options to integrate datasets for comparison: (1) BioRED+All (w/o AIONER): the model trained on the naive concatenation of all training datasets. (2) BioRED+All (MTL): the multi-task learning model trained on all datasets, in which each dataset is treated as an individual task. Table 2 shows the results of evaluating the models on the BioRED test set.

Compared to the baseline, the models trained on both the individual external datasets and BioRED training set can obtain better performance on the corresponding entity type with slightly higher F1-scores in overall performance. The MTL and AIONER models both perform significantly better than the baseline when all datasets are used for training. Particularly, the AIONER model obtained the highest overall score, improving the F1-score from 89.34% to 91.26%. In terms of entity types, disease and chemical are the most improved (4.60% and 2.43%, respectively). We evaluated the stability and robustness of AIONER and MTL models by comparing the means and standard deviations of their overall F-scores across 5 runs with different random initial seeds. Our results demonstrate that AIONER achieved a higher mean F-score and lower standard deviation than MTL, indicating its superior stability and robustness. Specifical-ly, the mean F-scores for AIONER and MTL were 91.21±0.15% and 90.64±0.34%, respectively. Compared with the MTL model that shares the same hidden layers but uses independent CRF output layers, our AIONER model merges all datasets with the AIO scheme and output results within a single output layer. Thus, it may be able to better utilize the information from the different datasets. Merging all datasets directly without applying the AIO scheme dropped the F1-score about 20% due to the large number of missing annotations. For example, gene datasets do not annotate diseases or chemicals.

**Table 2.** F1 scores for multiple named entity recognition on the BioRED test set

| Dataset | Overall | Gene | Disease | Chemical | Species | Variant | CellLine |
|---|---|---|---|---|---|---|---|
| BioRED | 89.34 | 92.35 | 83.47 | 88.55 | 96.98 | 87.34 | 90.53 |
| + NLM-Gene | 89.76 | 92.40 | 84.03 | 90.19 | 97.35 | 85.89 | 86.87 |
| + GNormPlus | 89.95 | **92.74** | 83.57 | 90.05 | 96.82 | 88.98 | 91.67 |
| + NCBI-Disease | 89.55 | 91.68 | 85.19 | 89.46 | 96.52 | 86.01 | 81.72 |
| + BC5CDR-Disease | 89.66 | 91.46 | 85.34 | 89.67 | 96.98 | 84.86 | 90.53 |
| + BC5CDR-Chemical | 89.40 | 91.52 | 84.07 | 89.09 | 96.99 | 88.38 | 87.50 |
| + NLM-Chem | 89.60 | 91.92 | 84.15 | 89.78 | 97.09 | 87.16 | 83.67 |
| + Linnaeus | 89.19 | 91.49 | 84.04 | 88.69 | 96.72 | 88.16 | 86.60 |
| + Species-800 | 89.65 | 92.19 | 83.34 | 90.14 | 97.37 | 88.79 | 80.81 |
| + tmVar3 | 89.01 | 91.08 | 83.77 | 88.09 | 97.08 | **89.21** | 88.66 |
| + BioID | 89.69 | 92.02 | 84.23 | 88.83 | 97.48 | 88.75 | **91.67** |
| + All (w/o AIONER) | 69.96 | 76.85 | 58.86 | 84.82 | 30.57 | 71.77 | 27.12 |
| + All (MTL) | 90.84* | 92.59 | 87.01 | 90.71 | 96.40 | 88.25 | 90.32 |
| + All (AIONER) | **91.26*** (+1.92) | 92.40 (+0.05) | **88.07** (+4.60) | **90.98** (+2.43) | **97.50** (+0.52) | 88.51 (+1.17) | 90.53 (+0.00) |

The parenthesized numbers are the improvements of AIONER compared to the baseline trained on the BioRED training set only. Bold indicates the best score for each entity type and overall entity. *p < 0.05 (two-sided Wilcoxon signed-rank test compared with baseline). There is no significant difference between MTL and AIONER.

Additionally, we tested the performance of the AIONER model trained on partial BioRED training data with all external datasets. We set up seven configurations with different numbers of abstracts (10, 50, 100, 200, 300, 400, and 500 abstracts) as the BioRED training subset. The results are shown in Figure 3. Here, the baseline is the model trained on the BioRED only. The results indicate that both AIONER and MTL models exhibit similar performances and outperform the baseline across all configurations of the BioRED training subset. Furthermore, it is noteworthy that even with only 10 articles, AIONER performs significantly better than MTL, and it still can achieve a competitive performance compared with the baseline model trained on the entire 500 articles (86.91% vs. 89.34%).

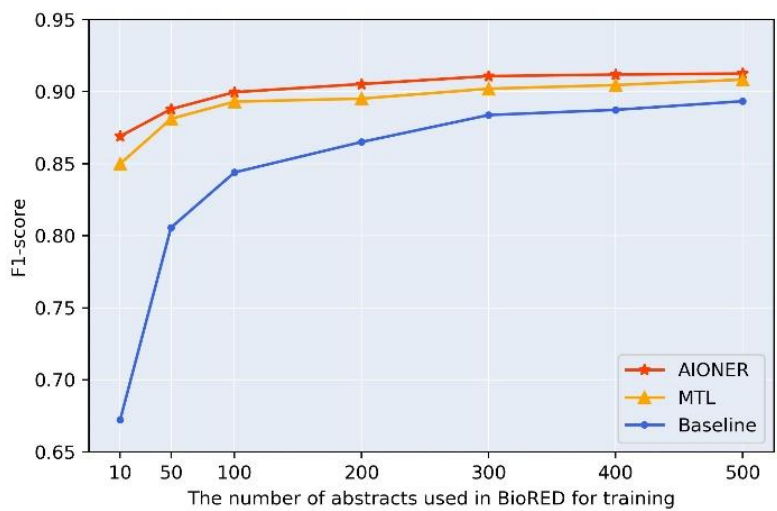

**Fig. 3. The performance of the models trained on different sizes of the BioRED training data.** Baseline: the model trained on the BioRED training set only; MTL: the multi-task learning model trained on the BioRED training set along with external datasets; AIONER: our AIONER model trained on the BioRED training set with external datasets.

### 3.3 Performance of AIONER on the test sets of individual datasets

The previous experiment demonstrated AIONER successfully utilized external datasets for the BioRED task. In this experiment, we evaluate the performance of the AIONER on those external datasets. We trained the PubMedBERT-CRF model only using the original training data as the baseline 1 (BL1). We provide an analysis of the performance of the model (baseline 2, BL2) on each individual corpus by training it on the combined dataset consisting of the individual corpus and the sub-corpus obtained by selecting the individual entity type from BioRED. The MTL model uses the same training data as AIONER.

As shown in Table 3, we obtained similar results to the previous exper-iment, both MTL and AIONER methods achieve higher average F1-scores than the baselines across those datasets. However, the perfor-mance of MTL is not stable and it performs worse than the baseline 1 on 3 out of 10 BioNER datasets. AIONER brings higher average improve-ments than MTL across those datasets. It performs better than the base-line on 9 out of 10 datasets and obtains significant advantages over the results of baselines and MTL on three datasets. Furthermore, our experi-ments revealed that the simple combination of individual corpus and sub-corpus from BioRED by selecting the corresponding entity type (i.e., baseline 2) resulted in an average F1-score lower than that of baseline 1. This suggests that the performance improvement cannot be achieved merely through the combination of the datasets. Notably, we observed a significant drop in the F1-score of baseline 2 on Linnaeus and Species-800 corpora, which could be attributed to the inherent differences in scope and quality between these corpora and BioRED for the species type. This highlights the challenges in leveraging existing datasets by combining them, especially when they differ significantly in scope and quality. In contrast, our proposed AIO schema improved the perfor-mance of datasets other than BioRED, demonstrating its high robustness and stability. We also provided a comparison with the state-of-the-art (SOTA) methods in terms of F1-scores for each corpus. AIONER achieved competitive performance compared to the SOTA methods. It is worth noting that a direct comparison with some published SOTA benchmarks for the BioID task may not be applicable due to the revised version of the BioID dataset in our experimental setup.

**Table 3.** F1 scores for the single entity recognition on the test sets of individual datasets

| Dataset | BL1 | BL2 | MTL | AIO | SOTA |
|---|---|---|---|---|---|
| NLM-Gene | 92.09 | 91.88 | 92.34 | **92.51** | 88.10 |
| GNormPlus | 85.09 | 85.92 | 85.62 | **85.98** | 86.70 |
| NCBI-Disease | 87.56 | 88.13 | 88.41 | **89.59*** | 89.71 |
| BC5CDR-Disease | 87.13 | 87.12 | 86.51 | **87.89*** | 87.28 |
| BC5CDR-Chemical | 93.42 | 92.82 | **93.93*** | 92.84 | 93.83 |
| NLM-Chem | 82.40 | 79.23 | **82.95** | 82.51 | 84.79 |
| Linnaeus | 90.36 | 85.19 | 90.14 | **90.63** | 92.70 |
| Species-800 | 78.32 | 76.91 | 78.76 | **79.67** | 76.35 |
| tmVar3 | 89.66 | 89.96 | 90.54 | **90.98** | 91.36 |
| BioID | 89.07 | 88.93 | 88.70 | **91.13*** | - |
| *Average* | 87.51 | 86.61 | 87.79 | **88.37** | - |

BL1: the PubMedBERT-CRF model trained on the original training set. BL2: the PubMedBERT-CRF model trained on a combination of individual corpora and sub-corpora decomposed from BioRED by selecting specific entity types. MTL: the multi-task learning model trained on the BioRED and the external datasets. AIO: the AIONER model trained on the BioRED and the external datasets. SOTA: the published state-of-the-art F1-score of each corpus. Bold denotes the best F1-score (except SOTA) on each dataset; *Denotes statistical significance over Baselines and MTL (two-sided Wilcoxon signed-rank test with a p-value < 0.05). We list the scores of the SOTA models on different datasets as follows: the score of Islamaj *et al.* (2021) on NLM-Gene, the score of Wei *et al.* (2019) on GNormPlus, the score of Zhang *et al.* (2021) on Species-800, the scores of Chai *et al.* (2022) on BC5CDR, the score of Tong *et al.* (2022) on NLM-Chem, the score of Sung *et al.* (2022) on Linnaeus, the score of Wei *et al.* (2022) on tmVar3. It is important to note that we made revisions to the BioID dataset in order to improve its consistency. As a result, certain previously published SOTA benchmarks for the BioID task (such as the F1-score of 74.4% reported in Arighi *et al.* (2017)) may not be directly comparable to our experimental setup.

**3.4 Using AIONER for new entity types**

Although our combined training dataset already covers six main entity types in the biomedical domain, there are other biological entities that are outside of our consideration. This experiment investigates whether AIONER can be applied to support those BioNER tasks with new entity types that are not previously seen in our AIONER training data. We conducted experiments with AIONER in two ways. (1) AIONER-Merged: we first merge the training set of the new task with all datasets via our AIO scheme to train a NER model, and then we applied the trained model to the task test set. (2) AIONER-Pretrained: We utilize the trained AIONER model as a pre-trained model, and then we further fine-tune this model on the new task with its training data. We benchmarked AIONER on three independent datasets with a variety of entity types different from the six which are our primary focus: (1) DMCB_Plant (Cho *et al.*, 2017) contains 3,985 mentions of plant names (e.g. "Trichosanthes kirilowii"). (2) AnEM (Pyysalo and Ananiadou, 2014) contains anatomical entities and organism parts between the molecule and the whole organism, with a total of 11 entity types (e.g. multi-tissue structure). (3) BEAR (Wührl and Klinger, 2022) annotates seven groups of biomedical entities (e.g., medical conditions, diagnostics, and environmental factors) on Twitter. More details of these datasets can be found in

Supplementary Table 2. We used the PubMedBERT-CRF model trained on the original training set as the baseline method. We also tested the MTL method for comparison. The overall performances of all models on the three individual tasks are shown in Figure 4.

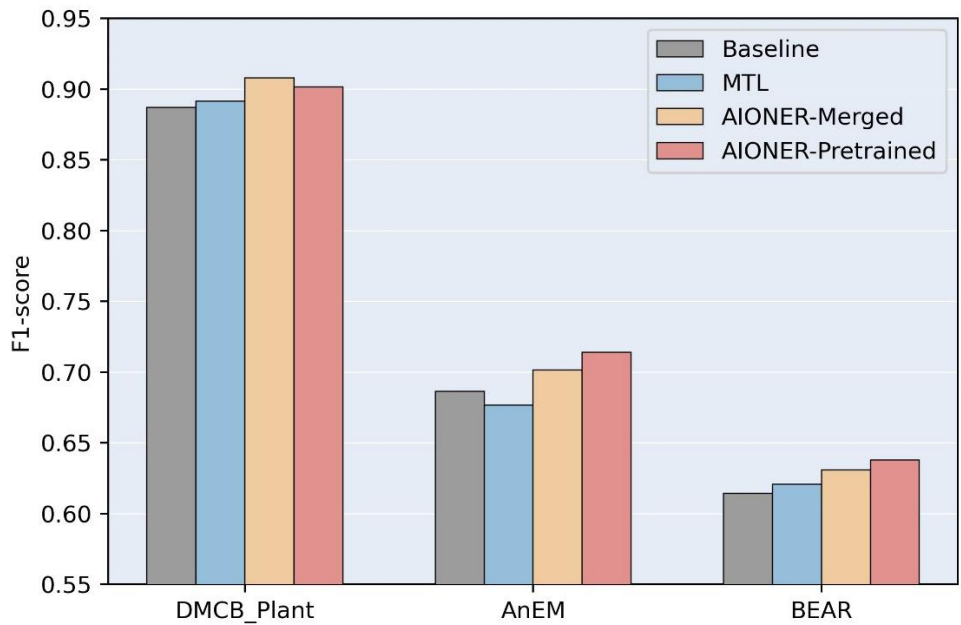

**Fig. 4. The performance of models on additional BioNER tasks.** Baseline: the model trained on the original training set. MTL: the multi-task learning model trained on the new dataset and all previous datasets. AIONER-Merged: the training set of the new task is first merged into all datasets via the AIO scheme, then the data is used to train the NER model. AIONER-Pretrained: the trained AIONER model is used as a pre-trained model, then the model is fine-tuned on the new task. AIONER-Pretrained significantly outperforms the baseline and MTL on AnEM; and it significantly outperforms the baseline on BEAR in a two-sided Wilcoxon signed-rank test with a p-value < 0.05.

Although none of the entity types in these three tasks are covered by BioRED, the dataset used to bridge between the individual datasets, AIONER can still improve the performance of the three tasks consistently. The results in Figure 4 indicate that the entity information gathered from the AIONER model assists in recognizing other entity types that have not been observed before. Overall, both the AIONER-Merged and -Pretrained models achieve better performance compared to the baseline and the MTL methods. More importantly, the performance gain is consistent for our AIONER method on all three tasks while MTL showed inferior performance on one of the three tasks. Again, this confirms that performance gain with MTL is somewhat task dependent, a limitation that was previously discussed in the literature (Chai *et al.*, 2022; Zhang and Yang, 2022).

**3.5 Performance of different deep learning model variants**

To process large-scale datasets in real-world settings, we further investigated the performance and efficiency of different deep learning models. The BioBERT (Lee *et al.*, 2020) and Bioformer (Fang and Wang, 2021) models were additionally evaluated as variants of BERT-based pre-trained language models (PLMs). For options of the output layer, we also tested the Softmax function to classify the label for each token. Table 4 shows the performance of different deep learning model variants on the BioRED test set.

**Table 4.** The performance of different deep learning model variants on the BioRED test set

| PLM | Output layer | Efficiency | | F1-score | |
|---|---|---|---|---|---|
| | | GPU | CPU | Baseline | AIO |
| PubMedBERT | CRF | 27s | 116s | **89.34** | **91.26** |
| | Softmax | 17s | 110s | 88.98 | 91.00 |
| BioBERT | CRF | 29s | 120s | 88.66 | 90.29 |
| | Softmax | 18s | 113s | 88.33 | 90.06 |
| Bioformer | CRF | 21s | 43s | 88.65 | 90.28 |
| | Softmax | **12s** | **40s** | 88.35 | 90.19 |

Baseline: the model trained on the original BioRED training set. AIO: the AIONER model trained on the merged training set. All AIONER models significantly outperform the corresponding baselines in a two-sided Wilcoxon signed-rank test with a p-value < 0.05. Bold indicates the best score in efficiency and F1-score. Note that the numbers of efficiency are the processing time (seconds) on the BioRED test set (100 abstracts). All models were evaluated on the same GPU (Tesla V100-SXM2-32GB) and CPU (Intel(R) Xeon(R) Gold 6226 CPU @ 2.70GHz, 24 Cores). The processing times of the BioBERT and PubMedBERT are almost the same, as their model architectures and parameters are similar.

The result shows that our AIONER scheme can be applied in various deep learning models and significantly enhances their performance. In comparison with other PLMs, the PubMedBERT model obtains the highest F1-scores. The lightweight Bioformer is more efficient and achieves a close performance. The efficiency advantage of the Bioformer has also been demonstrated in several recent studies (Fang and Wang, 2021; Luo *et al.*, 2022b). According to the statistical significance analysis, all PubMedBERT models exhibit significant improvements in performance compared to their corresponding BioBERT and Bioformer models. This finding is support-ed by a two-sided Wilcoxon signed-rank test with a p-value of less than 0.05. On the other hand, there was no significant difference observed between BioBERT and Bioformer models. In addition, the configuration of using CRF as the decoding layer obtains the best performance. The Softmax layer's performance is slightly lower, but the efficiency is significantly higher on GPU. No significant difference was observed between the performance of the same pre-trained language models with either the CRF or Softmax output layer. In summary, Bioformer-Softmax presents the highest efficiency (2- and 3-times improvement on both GPU and CPU servers respectively) and is very close to the best setting (PubMedBERT-CRF) in performance (about 1% drop in F-score). Moreover, Bioformer-Softmax-AIONER achieves higher performance than PubMedBERT-CRF-Baseline, suggesting it may be a better option for processing the large-scale datasets. Per our previous experience of extracting entities in entire PubMed abstracts (>35 Million) and PMC full texts (>4 Million) in PubTator Central (Wei *et al.*, 2019) by individual entity taggers, the whole process took ~30 days by using 300 parallel processes via NCBI computer cluster. As an estimation of using Bioformer-Softmax-AIONER instead of the six individual NER tools, the processing time can be reduced to less than 10 days. Thus, the implementation of the AIONER brings significant advantages of entity recognition on large-scale data.

## 4 Discussion

As mentioned in the "Introduction" section, several MTL methods have been explored for BioNER tasks to make full use of various existing resources. These methods share the hidden layers to learn common features, which are generic and invariant to all the tasks. MTL is powerful when all the tasks are related, but it is vulnerable to noisy and outlier tasks, which can significantly degrade performance (Zhang and Yang, 2022). Different from MTL, our AIONER scheme successfully integrates multiple resources into a single task via adding task-oriented tagging labels. The learned features are more informative and flexible. The results of our experiments demonstrate our AIONER method achieved performance competitive with the MTL methods for multiple named entity recognition, and it is more stable than the MTL method on multiple BioNER tasks. Moreover, AIONER does not require complex model design and it can be easily implemented with various machine learning models.

The main contribution of the AIONER schema is that it can train on more diverse entity-type corpora. By doing so, the model can learn dif-ferent synonyms present in diverse texts. We analyzed the differences between the results of the AIONER model trained on multiple corpora and the baseline model trained on the BioRED corpus only. Then, we summarized the three main cases where AIONER performs better than the baseline. (1) More precise categorization of entity types: AIONER can better categorize the entity type based on the context. In PMID:15464247, "HCV genotype 1-infected" is incorrectly recognized as a species by the baseline, but AIONER can correctly detect it as a disease. (2) Better boundary detection: AIONER also presents higher accuracy in detecting the entity boundaries. For example, in the case of "Vogt-Koyanagi-Harada (VKH) syndrome", the baseline detects the wrong boundaries and wrongly identifies it as two entities, "Vogt-Koyanagi-Harada" and "(VKH) syndrome," while AIONER correctly recognizes it. (3) Slightly higher recall: Some entity spans that are not shown in the training set may be missed by the baseline, but AIONER can handle those unseen entities better.

Although AIONER exhibits promising performance for multiple named entity recognition, there are still some errors in the results. We have reviewed the errors of the model with the best performance (BioRED + All via AIONER) on the BioRED test set and sorted the errors based on the percentages as shown in Figure 5. (1) Incorrect boundary (36.0%): most errors are caused by incorrect boundaries in the extracted mentions. In this error type, the most critical issue is which leading or trailing tokens fall within the mention boundary. For example, "Necrotising fasciitis" (MeSH: D019115) is the disease mention that should be detected, but our method missed the first token "Necrotising" which can help to narrow down the entity more specifically. (2) Entity type ambiguity (26.6%): This error type contains two major errors. First, the same entity mention may have different entity types in the text. For example, Growth hormone is a protein encoded by the gene that is a member of the somatotropin/prolactin family which plays an important role in growth control and is also a drug (chemical) for the treatment of the growth hormone deficiency. In different context, it can be annotated differently. Second, different entities may be ambiguously named. For example, BMD gene is the corresponding gene of the Becker Muscular Dystrophy (BMD). Both the gene and the disease are named BMD. (3) Natural language mentions (10.4%): Pre-trained language models are trained on large natural language texts. However, it is still very difficult to accurately identify the entities written in descriptive natural language

(e.g., the variant of "phenylalanine to the polar hydrophilic cysteine in exon 6 at codon 482"). Similarly, composite spans – which mention multiple entities – also confuse the models (e.g., D1 or D2 dopamine receptors). Such natural language mentions are rare in the training set and individually unique, making recognition challenging. Other smaller error types include missed entities, which are mostly abbreviations with insufficient definition, entities not in the recognition scope which were wrongly detected, multiple spans of the same entity were detected inconsistently, and others. Our future work will focus on these problems, incorporating linguistic information (e.g., part of speech and syntactic information) and dictionary resources into our method to further enhance the model's performance.

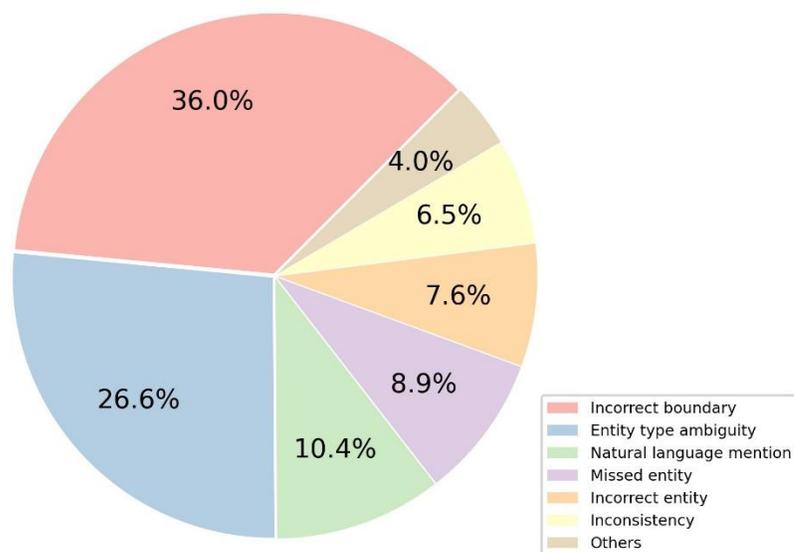

**Fig. 5.** Error analysis of the AIONER results on the BioRED test set.

AIONER is a reliable method for recognizing the entities of different types at once. However, AIONER cannot return multiple entities with overlapping boundaries, such as a nested entity span. For example, growth hormone deficiency contains two mentions, "growth hormone" and "growth hormone deficiency." AIONER cannot return both simultaneously.

## 5 Conclusion

In conclusion, we present an AIONER method to integrate heterogeneous corpora for multiple named entity recognition at once. AIONER can develop a single model for multiple entity types with optimized performance for generalizable usage. This implementation also significantly reduces the effort of the process, especially for large-scale data. We also demonstrate that AIONER can be used to further improve the performance of various BioNER tasks, even when the entity types have never been observed before. These results suggest that AIONER is highly robust and generalizable for BioNER. We released the pre-trained AIONER model for standalone usage to support the BioNER tasks. In the future, we will apply the optimized AIONER model on the entire PubMed (abstracts) and PMC full texts for downstream text mining research (e.g., biomedical relation extraction).

## Funding

This research was supported by the Intramural Research Program of the National Library of Medicine (NLM), National Institutes of Health; the Fundamental Research Funds for the Central Universities [DUT23RC(3)014 to L.L.]

*Conflict of Interest:* none declared.

# (Supplementary Information)

**Supplementary Table 1. The existing datasets within the six popular concepts in biomedical literature.**

| Dataset | Gene | Disease | Chemical | Variant | Species | CellLine |
|---|---|---|---|---|---|---|
| BioRED (Luo *et al.*, 2022) | ○ | ○ | ○ | ○ | ○ | ○ |
| CRAFT (Bada *et al.*, 2012) | ○ | | ○ | | | ○ |
| GNormPlus (Wei *et al.*, 2015) | ○ | | | | | |
| NLM-Gene (Islamaj *et al.*, 2021) | ○ | | | | | |
| BioCreative II GM (Smith *et al.*, 2008) | ○ | | | | | |
| ChemProt (Krallinger *et al.*, 2017) | ○ | | ○ | | | |
| DrugProt (Miranda *et al.*, 2021) | ○ | | ○ | | | |
| JNLPBA (Kim *et al.*, 2004) | ○ | | | | | ○ |
| GPRO (Krallinger *et al.*, 2015) | ○ | | | | | |
| RENET (Wu *et al.*, 2019) | ○ | ○ | | | | |
| RENET2 (Su *et al.*, 2021) | ○ | ○ | | | | |
| NCBI Disease (Doğan *et al.*, 2014) | | ○ | | | | |
| BC5CDR (Li *et al.*, 2016) | | ○ | ○ | | | |
| NLM-Chem (Islamaj *et al.*, 2021) | | | ○ | | | |
| CHEMDNER (Krallinger *et al.*, 2015) | | | ○ | | | |
| CEMP (Krallinger *et al.*, 2015) | | | ○ | | | |
| tmVar1-3 (Wei *et al.*, 2022; Wei *et al.*, 2013; Wei *et al.*, 2018) | | | | ○ | | |
| Nala (Cejuela *et al.*, 2017) | | | | ○ | | |
| SETH (Thomas *et al.*, 2016) | | | | ○ | | |
| OSIRIS (Furlong *et al.*, 2008) | | | | ○ | | |
| Linnaeus (Gerner *et al.*, 2010) | | | | | ○ | |
| SPECIES (Pafilis *et al.*, 2013) | | | | | ○ | |
| BioID corpus (Arighi *et al.*, 2017) | | | | | | ○ |
| CellFinder (Kaewphan *et al.*, 2016) | | | | | | ○ |

**Supplementary Table 2. The corpora for the evaluation of BioNER tasks with new entity types.**

| Task | # Entity type | Text genre | Text size | # Entity |
|---|---|---|---|---|
| DMCB_Plant (Cho *et al.*, 2017) | 1 | PubMed abstract | 208 | 3,985 |
| AnEM (Pyysalo and Ananiadou, 2014) | 11 | PubMed abstract + PMC full-text | 500 | 3,135 |
| BEAR (Wührl and Klinger, 2022) | 14 | Twitter | 2,100 | 6,324 |

**Table S3. Comparison of different task-specific tagging modes on the BioRED test set.** We examine the task-specific tagging modes for identifying all entities in the BioRED test set. The all-in-one (AIO) mode employs the "<ALL></ALL>" to directly recognize all concept entities. On the other hand, the individual (IND) mode predicts the corresponding entity type by using each individual tag (e.g., <Gene></Gene> for gene type) and then unified the results to compute the performance. When the predicted entities overlap, we select the longest entity as the final result. To investigate if the performance of the IND and AIO modes can be improved by combining their results, we integrated the output of both modes and included the combined results in the last row of the table (denoted as "Combined"). The table presented below illustrates the results, indicating that the AIO mode achieves higher F1-scores for overall performance and each entity type on BioRED. This is primarily because the entity scope and definition in individual corpora do not align completely with BioRED. For instance, Linnaeus and Species-800 do not annotate species relevant clinical terms, such as "patient". Combining all predicted results of IND and AIO did not improve the performance, leading us to recommend using the AIO tag mode to identify all concept entities in BioRED and the IND marks to recognize the entities in the individual corpora.

| Task-specific tagging mode | Overall | Gene | Disease | Chemical | Species | Variant | CellLine |
|---|---|---|---|---|---|---|---|
| IND mode | 86.37 | 91.26 | 87.27 | 88.08 | 57.24 | 88.02 | 88.17 |
| AIO mode | **91.26** | **92.40** | **88.07** | **90.98** | **97.50** | **88.51** | **90.53** |
| Combined | 90.42 | 91.73 | 87.28 | 89.48 | 97.50 | 87.37 | 90.32 |

**Experimental Setup and Hyper-parameter Tuning**

In terms of hyperparameters, we focused on optimizing the learning rate, and did not perform any additional optimization. Specifically, we randomly selected 10% of the training set as a development set to determine the best learning rate, selecting from a set of four possible options: 1e-5, 5e-5, 5e-6, and 1e-6. The learning rate was chosen based on the highest F1-score on the development set, and the development set was then combined with the training set for subsequent model evaluation. The number of training epochs was determined using the patience parameter, which was set to 5, with a maximum of 50 epochs. The training was stopped if there was no improvement in accuracy after five consecutive epochs. To ensure fairness in comparison, each developed model was optimized using this method.

The primary hyperparameters of AIONER were set as follows: a learning rate of 5e-6, batch size of 32, and a maximum input length of 256 tokens. Our NER models take sentences as input, and based on the training data we used, only 0.04% of sentences were longer than 256 tokens. Setting the maximum input length to 256 tokens allows the model to effectively cover most sentence lengths, and helps to ensure more efficient training and testing. We also tested a maximum input length of 512 tokens, but found that PubMedBERT did not perform better than the model with a length of 256 tokens (91.01% vs. 91.26%).